%% file: lrec-coling2024-example.tex
\documentclass[10pt, a4paper]{article}

\usepackage{lrec-coling2024} 

\usepackage{natbib}
\usepackage{multibib}
\makeatletter
\def\@mb@citenamelist{cite,citep,citet,citealp,citealt,citepalias,citetalias}
\makeatother
\newcites{languageresource}{~}

\usepackage{graphicx}
\usepackage{tabularx}
\usepackage{soul}
\usepackage{titlesec}
\titleformat{\section}{\normalfont\large\bfseries\center}{\thesection.}{1em}{}
\titleformat{\subsection}{\normalfont\SmallTitleFont\bfseries\raggedright}{\thesubsection.}{1em}{}
\titleformat{\subsubsection}{\normalfont\normalsize\bfseries\raggedright}{\thesubsubsection.}{1em}{}
\renewcommand\thesection{\arabic{section}}
\renewcommand\thesubsection{\thesection.\arabic{subsection}}
\renewcommand\thesubsubsection{\thesubsection.\arabic{subsubsection}}

\usepackage{xcolor}
\usepackage{hyperref}
 \definecolor{darkblue}{rgb}{0, 0, 0.5}
  \hypersetup{colorlinks=true, citecolor=darkblue, linkcolor=darkblue, urlcolor=darkblue}

\usepackage{xstring}

\usepackage{color}
\usepackage{amsmath}
\usepackage{booktabs}
\usepackage{multirow}

\title{UniRetriever: Multi-task Candidates Selection for Various \\ Context-Adaptive Conversational Retrieval}

\name{Hongru Wang$^{\heartsuit \nabla}$, Boyang Xue$^{\heartsuit \nabla}$, Baohang Zhou$^{\spadesuit}$, Rui Wang$^{\clubsuit}$, \\ {\bf \large Fei Mi$^{\diamondsuit}$, Weichao Wang$^{\diamondsuit}$, Yasheng Wang$^{\diamondsuit}$, Kam-fai Wong$^{\heartsuit \nabla}$}}

\address{$^{\heartsuit}$ MoE Key Laboratory of High Confidence Software Technologies, China \\ 
    $^{\nabla}$ The Chinese University of Hong Kong
    $^{\spadesuit}$ Nankai University  \\
    $^{\clubsuit}$ Harbin Institute of Technology, Shenzhen
    $^{\diamondsuit}$ Huawei Noah's Ark Lab \\
    \{hrwang, byxue\}@se.cuhk.edu.hk, zhoubaohang@dbis.nankai.edu.cn, ruiwangnlp@outlook.com \\}

\abstract{
Conversational retrieval refers to an information retrieval system that operates in an iterative and interactive manner, requiring the retrieval of various external resources, such as persona, knowledge, and even response, to effectively engage with the user and successfully complete the dialogue. However, most previous work trained independent retrievers for each specific resource, resulting in sub-optimal performance and low efficiency. Thus, we propose a multi-task framework function as a universal retriever for three dominant retrieval tasks during the conversation: persona selection, knowledge selection, and response selection. To this end, we design a dual-encoder architecture consisting of a context-adaptive dialogue encoder and a candidate encoder, aiming to attention to the relevant context from the long dialogue and retrieve suitable candidates by simply a dot product. Furthermore, we introduce two loss constraints to capture the subtle relationship between dialogue context and different candidates by regarding historically selected candidates as hard negatives. Extensive experiments and analysis establish state-of-the-art retrieval quality both within and outside its training domain, revealing the promising potential and generalization capability of our model to serve as a universal retriever for different candidate selection tasks simultaneously.
 \\ \newline \Keywords{Conversational Retrieval, Knowledge Selection, Response Selection, Multi-task Framework} }

\begin{document}

\maketitleabstract

\section{Introduction}
\input{figs/intro.tex}


Information Retrieval (IR) refers to the task of retrieving the most relevant candidates (e.g. top \textit{n}) from a large corpus (a.k.a, candidates pool) for a given query, which receives a rapid proliferation of interest and attention in both academia and industry \cite{DBLP:conf/iclr/IzacardG21, ernie-search, chen-etal-2022-ketod}. These retrieved evidence serve as additional semantic signals to provide important information, guiding the generation of the final answer in many downstream natural language tasks such as question answering \cite{dpr, izacard-grave-2021-leveraging, rankingT5}, machine reading comprehension \cite{colbert, rocketqav2,qu-etal-2021-rocketqa} and also dialogue systems \cite{wow,zhou-etal-2020-kdconv,dulemon,wan-etal-2023-multi}. Benefiting from more accurate retrieved results, it is observed and well-acknowledged that the performance of these downstream tasks can be further improved \citep{wow, shuster-etal-2022-language, dulemon} with the more powerful retriever such as DPR \citep{dpr}, and RocketQA \citep{qu-etal-2021-rocketqa, rocketqav2}.



Unlike traditional Information Retrieval (IR) systems, Conversational Retrieval (CR) is an embodiment of an iterative and interactive IR system that has two distinct characteristics. On the one hand, the dialogue context is much longer than the question (a.k.a, the query) in the question-answering field \citep{dpr,izacard-grave-2021-leveraging} because dialogue can last for many sessions \cite{orcqa}, as seen in the Multi-Session Chat dataset \cite{msc}, proposing additional challenges and difficulty in locating relevant contextual information and modeling the relationship between the query and the candidate. 
Despite the query rewriting \cite{wu-etal-2022-conqrr} is a possible way to tackle the lengthy input, it necessitates large amounts of labeled data and still requires locating the relevant contextual turns once the length of multi-session dialogue exceeds the maximum input limit of the language models \citep{bert,rankingT5,llama}. Furthermore, in the ongoing conversation, multiple turns may revolve around a common topic but draw upon various external candidates. The subtle differences between these candidates play a key role in determining the order relationship between them and the current context.

On the other hand, various external candidates\footnote{To avoid ambiguity, we use ``candidates" to represent all required sources in dialogue such as ``persona", ``knowledge" and others.}, such as persona and knowledge sources, need to be retrieved to engage the users and complete the dialogue goal during the interactions \citep{wang2023survey, wang2024unimsrag}. For example, a human-like dialogue system not only needs to retrieve suitable persona descriptions \cite{dulemon,persona_extending} to maintain a consistent personality but also external knowledge such as Wikipedia\footnote{https://www.wikipedia.org/} to answer the user's query \cite{wow}. These various candidates serve as crucial plugins to enhance the quality of responses, ensuring they are personalized, informative, coherent, and encompassing other vital features, depending on the particular source employed \citep{wang-etal-2023-large, wang2023tpe}. However, previous works usually train an independent retriever for each source, resulting in sub-optimal performance and inefficient computing \citep{wow,dulemon}.



In response to these problems, we propose the \textbf{Universal} \textbf{C}onversational \textbf{R}etrieval (a.k.a. UniversalCR), a multi-task framework to advocate for a unifying view of three dominant candidates selection tasks in the conversation, including \textit{persona selection}, \textit{knowledge selection} and \textit{response selection}. As shown in Figure~\ref{intro_example}, to respond to the final turn $U_3$, the dialogue system needs to select relevant persona $P_n$ and knowledge $K_n$ when generating the response similar to $R_2$. Besides that, the response selection task has always been a hot research topic in the retrieval-based dialogue systems \citep{DBLP:conf/cikm/HuaFTY020,gu-etal-2019-dually,DBLP:conf/sigir/GuLLLCZ21}. It is observed that the persona $P_n$, knowledge $K_n$, and the ground-truth response $R_2$ share similar semantics, which motivates us to consider approaching these candidate selection tasks using a multi-task approach. Specifically, we first design a context-adaptive dialogue encoder to dynamically select related dialogue histories according to the query (i.e. the last utterance in the dialogue), while discarding noisy and unrelated utterances from lengthy dialogues. Then, we utilize historically selected candidates ($P_2$, $K_1$ and $R_1$ in the Figure~\ref{intro_example} for current turn) during conversation to propose \textit{historical contrastive loss}, while regarding historically selected candidates as \textbf{semi-hard negatives} and randomly sampled candidates as \textbf{easy negatives}. In addition, we utilize \textit{pairwise similarity loss} to rank different pairs of candidates and dialogue context, inspired by previous negative sample mining \citep{xuan2020hard,zhou2022simans}. To summarize, this work makes the following contributions:








\begin{itemize}
    \item We propose a universal conversational retrieval framework, unifying three dominant candidate selection tasks: \textit{persona selection}, \textit{knowledge selection}, and \textit{response selection}, in one framework while keeping the bottleneck layer as a single dot-product with a fixed size to achieve the balance of effectiveness and efficiency.

    \item We design one context-adaptive encoder and two carefully crafted loss constraints to address lengthy dialogue and capture subtle differences across various candidates respectively.

    \item We conduct extensive experiments to demonstrate the superiority of our proposed framework on six datasets in both supervised and unsupervised settings. Besides that, we offer an in-depth analysis of various candidate pool sizes and different context processing methods. These findings suggest a promising path toward building a robust and universal dialogue retrieval framework.
\end{itemize}

\section{Related Work}
\label{sec:related_work}

\noindent \textbf{Conversational Retrieval. }  Information Retrieval (IR) has been investigated and used in many applications such as web search and digital libraries, aiming to retrieve a ranked list of relevant documents in response to the query, while conversational retrieval is one embodied IR system. Most previous works conduct conversational retrieval in the context of conversational question answering, following the retrieval-then-rank framework in the traditional IR systems \cite{orcqa,wu-etal-2021-dialki,rerank_overgenerated_response,dai2022dialoginpainting}. However, they always are fine-tuned for a specific type of resource and thus limited in application and generalization \cite{ConvDR,kim-kim-2022-saving}. For example, \citet{wu-etal-2021-dialki} introduces a knowledge identification model with an auxiliary task that predicts previously used knowledge to capture the history of dialogue-document connections. Instead, we directly regard previously used candidates\footnote{It could be anything, here we denote persona, knowledge, and response.} as the hard negative sample to model the relationships among different candidates. Additionally, many researchers have proposed different methods to fine-tune dual encoder retrievers \cite{realm,dpr,lin-etal-2021-contextualized,kim-kim-2022-saving},  such as coupling both coarse retrieval and fine reranking features to facilitate the final retrieval performance \cite{kumar-callan-2020-making,HLATR}. 

Recently, due to the exceptional performance of large language model (LLM) on various downstream tasks \citep{bang-etal-2023-multitask}, there are some attempts which directly prompt LLMs to function as a retriever or re-ranker \citep{zhu2024large, wang2024unimsrag}. For example, \citet{sun-etal-2023-chatgpt} evaluate the performance of LLMs as a re-ranker, and a very recent work formulate conversational retrieval as relevance score prediction task, which is optimized with knowledge source selection and response generation in a multi-task manner using a single LLM \citep{wang2024unimsrag}. It is worthy noting  retrievers, rather than re-rankers, are typically applied to thousands of documents or queries, posing inefficiency and affordability challenges when using LLMs (no matter using fine-tuning or prompting). Additionally, LLMs are not optimal solutions in certain cases \citep{ma2023zeroshot, wang2024unimsrag}.

\noindent \textbf{Pre-trained LMs for Dialogue. } Large language models such as ToDBERT \cite{todbert} and DialoGPT \cite{zhang-etal-2020-dialogpt} have shown impressive open-ended capabilities in both understanding and generation tasks after in-domain per-training or fine-tuning. \citet{dialoguebert} propose a novel contextual dialogue encoder (i.e. DialogueBERT) with five well-designed pre-training tasks including response selection. Besides that, there are also some works that design specific architectures and embeddings to effectively exploit the semantic information in dialogues \cite{hae_bert,gu2021dialogbert}. Our work also is in line with these previous works, by training a universal and special retrieval for dialogue, specifically long ones, to retrieve various external candidates needed to engage users and complete the dialogue goal.

\section{Method}

In this section, we will introduce the three modules of our proposed framework one by one: context-adaptive encoder, candidate encoder, and the final training objectives, starting with a formal problem definition. The whole framework is illustrated in Figure~\ref{model}.

\subsection{Problem Definitions}
Given a long dialogue $\mathcal{D}_{T}=\{(u_i, s_i)_{i=0}^T\}$, $u_i$ and $s_i$ are $i_{th}$ user utterance and system utterance respectively, for current dialogue context $\mathcal{D}_{context}$ consisting of $\mathcal{D}_{T-1}$ and $u_t$, the model is required to retrieve appropriate persona $p$ (persona selection), knowledge $k$ (knowledge selection), and sometimes even $s_t$ itself (response selection) from a corresponding candidate pool to accomplish the dialogue and engage the users,

\begin{equation}
    c_{*} = \arg \max_{c \in C} \text{sim}(q, c)
\end{equation}

where $C$ is the candidate pool, $q$ is the query consisting of $(D_{T-1}, u_t)$, $c$ is a candidate from the pool which is composed of a sequence of words, and $\text{sim}(q, c)$ is a similarity function\footnote{Without other statements, we adapt a simple dot product as the similarity function by default.} that measures the similarity between the query and candidate. The objective of this function is to find the candidate $c$ in the pool $C$ that has the highest similarity score with the query $q$. Here $c$ can either be persona, knowledge, or response.

\subsection{Context-adaptive Encoder}
To locate the relevant contextual information in the lengthy dialogues, we choose to process individual utterances ($u_i$ or $s_i$) through the encoder instead of combining them all into a single input, which often exceeds the maximum input limit or introduces unnecessary noises. In detail, we feed the utterance into the encoder with special indicator tokens \texttt{[CLS]} and \texttt{[USR]} or \texttt{[SYS]} at the beginning to obtain its representation \footnote{To exploit discourse-level coherence among utterances \cite{gu2021dialogbert}, we can add the 2-d position embedding based on the order of utterance in dialogue (discourse level) and order of word in utterance (utterance level).}:

\begin{equation}
\label{enc}
    h_i = \mathbf{Enc} (utter)
\end{equation}

where $utter \in \{u_i, s_i\}$ and $h_i \in \{h^u_i, h^s_i\}$ accordingly. Considering that the dialogue may span multiple sessions \cite{msc}\footnote{If not, we can simply regard one turn $[u_i, s_i]$ as one session unit.}, we first divide the given long dialogue into the previous sessions $D_{prev}$ and the current session $D_{curr}$. Inspired by lots of previous work which proved that the last utterance $h^u_t$ plays a key role in retrieving relevant resource \cite{dulemon}, here we regard it as the query to retrieve relevant utterance in the previous session:

\input{figs/model.tex}

\begin{equation}
\label{k_ret}
    H_{prev} = TopK_{h_j \in D_{prev}} ( sim(h^u_t, h_j) )
\end{equation}

Thus we can easily filter unrelated and redundant utterances in the previous session\footnote{It is well noted that the retriever becomes more accurate since the encoder gets optimized as the training continues.}. Then we concatenate retrieved $H_{prev}$ with $H_{curr}$ from the current session $[H_{prev}; H_{curr}]$ to form $H_{hist}$ as key and value to learn contextualized embeddings of multi-turn dialogues, here $H_{curr}$ consist of utterances from the current session except the last one.

\begin{equation}
    h_{hist} = \mathbf{Attn}(u_t, H_{hist}, H_{hist})
\end{equation}

In order to control the query and dialogue history contribution in the final representation, we add a gate after the attention block,

\begin{equation}
\begin{split}
    h_{d} &= \lambda * h_{hist} + (1 - \lambda) * h^u_{t} \\
    \lambda &= \sigma(\mathbf{w} * [h_{hist};h^u_t])
\end{split}
\end{equation}

where $\mathbf{w}$, $\sigma$, $h_d$ indicate a learnable parameter, sigmoid function and the final context representation respectively.

\subsection{Candidates Encoder}
To unify different candidate selection tasks into one framework, we design unique tokens to indicate each task. Specifically, we use \texttt{[PERSONA]} for persona selection, \texttt{[KNOWLEDGE]} for knowledge selection, and \texttt{[RESPONSE]} for response selection. Thus the model can easily recognize each task and perform the corresponding retrieval seamlessly. Then we feed it to the same encoder described in Eq.\ref{enc}

\begin{equation}
    c_i = \mathbf{Enc} (cand)
\end{equation}

Where $cand$ consists of \texttt{[CLS]} \texttt{[CANDIDATES]} $w_1, w_2, ..., w_n$ in which \texttt{[CANDIDATES]} can be replaced by any candidate selection task indicator token described above. Thus, the framework can be easily extended to other candidate selection tasks and activated by specific candidate tokens without the necessity of training separate retrievers for each individual candidate selection task.

\subsection{Training Objectives}
Negative sample mining is vital to effectively train the dense retrieval model \cite{hard_negative_mining}. Previous work empirically showed the negatives ranked around the positives are generally more informative and less likely to be false negatives, requiring more attention \cite{zhou2022simans}. Building on these previous findings, we introduce a novel approach by considering the historically selected candidates as semi-hard negatives, which share closer semantics to the positives (as showed in Figure~\ref{intro_example}) than random negatives, to design two distinct loss objectives, namely \textit{historical contrastive learning} and \textit{pairwise similarity loss}.

\bigskip
\noindent \textbf{Historical Contrastive Learning. }
Instead of predicting historically selected candidates \cite{wu-etal-2021-dialki}, we take advantage of similar but different semantics between historical candidates and current ones by regarding the former as \textit{\textbf{semi-hard negative samples}}. For example, for the dialogue context in Figure~\ref{intro_example}, the $P_n$ is the positive persona for the current turn, $P_2$ is the semi-hard negative, and other personas such as $P_1$ is the randomly sampled easy negative. With semi-hard negative samples in the batch, the model is optimized by:

\begin{equation}
\mathcal{L}_{hist} = \frac{e^{sim(h_d^i, c_i^{+})}}{\sum_{j\in \mathcal{B}}{e^{sim(h_d^i,c_j^{+})} + e^{sim(h_d^i, c_i^{-})}}}
\end{equation}

where \textit{sim} is a similarity function, $\mathcal{B}$ is a mini-batch of examples, $c_i^+$ and $c_i^-$ are positive candidates and semi-hard negative candidates for $i_{th}$ dialogue context $h_d^i$. Once there are no semi-hard negatives or the semi-hard negative is the same as the positive, we directly use randomly sampled easy negative as $c_i^-$. In this way, the loss function simplifies to a conventional contrastive loss. By including historical negative candidates in the training data, the model is forced to learn to identify the subtle and key information required for the current turn instead of useless or redundant ones.

\bigskip
\noindent \textbf{Pair-wise Similarity Loss. }
Instead of considering each context-candidate pair in isolation, we can alternatively focus on pairwise comparisons \cite{rankingT5} to improve the accuracy of our ranking. To achieve this goal, we trained our model using a modified pairwise circle loss function \cite{circle_loss}. This loss function has a unified formula that can be used for two fundamental paradigms in deep feature learning: learning with class-level labels and learning with pairwise labels. The original loss function is shown below:

\begin{equation}
\mathcal{L}_{pair} = log [1 + \sum_{i=1}^{K}\sum_{j=1}^{L} e^{\gamma (s_n^j - s_p^i)} ]
\end{equation}

where $\gamma$ is a scale factor $s_n$ and $s_p$ stands for two pairs where $s_n < s_p$. Specifically, here we have three different pairs as shown in Figure \ref{model}: \textit{(context, positive candidate), (context, historical candidate)} and \textit{(context, negative candidate)}. As such, we can have the following preference ranking $r_{pos} > r_{hist} > r_{neg}$. Then the modified loss objective becomes:


\begin{equation}
\begin{split}
\mathcal{L}_{pair} &= log[1 + \sum_{i=1}^K e^{\gamma (s_{neg}^i - 
         s_{hist}^i)} \\
    & + \sum_{j=1}^L e^{\gamma (s_{hist}^j - s_{pos}^j)}]
\end{split}
\label{pairsim_loss}
\end{equation}

Where $K$ and $L$ denote the number of similarity scores, $s_{neg}$ denotes the similarity between context and the negative candidate, and so on\footnote{We found add $r_{pos} > r_{neg}$ order relationship in Eq.~\ref{pairsim_loss} explicitly can not bring significant improvement.}. Thus, the partial-order relationship between dialogue context and different candidates can be modeled and captured, which is the key point of a re-ranking module \cite{HLATR}. 

Lastly, we combine these two losses at the same time to form the final training objective.


\begin{equation}
\label{loss_obj}
\mathcal{L} = \mathcal{L}_{hist} + \mathcal{L}_{pair}
\end{equation}

\section{Experiments}
In this section, we first introduce our six used datasets for persona selection, knowledge selection, and response selection tasks, and then present baselines and our main experimental results.

\input{tables/data_sts.tex}

\input{tables/main_experiments.tex}

\subsection{Datasets}

\textbf{DuLeMon. }\cite{dulemon} a latest open-domain dialogue dataset with long-term persona memory in which a response is grounded on persona information that occurred in historical sessions, leading to better dialogue engagingness. \textbf{\textit{persona selection}}

\textbf{Knowledge Behind Persona (aka, KBP)}. a dialogue dataset, in which the response is grounded on both the persona and its corresponding implicit knowledge \citep{wang-etal-2023-large}. We utilize this data to evaluate the out-of-domain and zero-shot performance of our model.  \textbf{\textit{persona selection}}

\textbf{KiDial-S. } another collected knowledge-grounded dialogue set following \citet{dai2022dialoginpainting}, which automatically turning knowledge documents into simulated multi-turn dialogues \citep{wang-etal-2023-retrieval}.  \textbf{\textit{knowledge selection}}

\textbf{DuSinc. }\cite{dusinc}  an open-domain human-human dialogue dataset, where a participant can access the service to get the information needed for dialogue responses. \textbf{\textit{knowledge selection}}
    
\textbf{Diamante. }\cite{diamante}, a chit-chat dataset by asking annotators to select or amend the model-generated candidate responses. Since the dataset contains human-generated responses and model-generated responses, we regard the former as positive samples, the latter as hard negative samples, and random responses as negative. \textbf{\textit{response selection}}
    
\textbf{KdConv. }\cite{zhou-etal-2020-kdconv} a multi-domain dialogue dataset towards the multi-turn knowledge-driven conversation. Since it grounds the response to knowledge graphs, we do not consider it for knowledge selection. \textbf{\textit{response selection}}

\subsection{Baselines}


To better evaluate the performance of our proposed method, we conducted extensive experiments with different baselines, including both sparse and dense retrieval models: (1) \textbf{BM25} \cite{bm25} (2) \textbf{DPR} \cite{dpr}
(3) \textbf{SentenceBERT} \cite{sentenceBERT}, (4) \textbf{PolyEncoder} \cite{polyencoder}, (5) \textbf{Bi-Encoder}, (6) \textbf{RocketQAv1} \cite{qu-etal-2021-rocketqa} and \textbf{RocketQAv2} \cite{rocketqav2} \footnote{https://github.com/PaddlePaddle/RocketQA}, (7) \textbf{MultiCPR} \cite{Long2022MultiCPRAM}, (8) \textbf{UniversalCR}$_{single}$, and (9) \textbf{UniversalCR}$_{full}$. For the former 8 models, we train them independently for each candidate selection task, and we only adopt multi-task learning for \textbf{UniversalCR}$_{full}$. We chose these baselines since they mostly implemented typical dual-encoder architecture while adapting different interaction strategies: late interaction such as PolyEncoder, hard negative mining such as RocketQAv2, and knowledge distillation such as MultiCPR, without fancy and complicated architecture, resulting in more efficient computing. In addition, there are some methods that take advantage of more than one strategy such as RockerQAv2. We emphasize that almost all of these strategies can be plugged into our framework. We left this in our future work.

\subsection{Implementation Details}
We utilize LERT-base \cite{cui2022lert} as the backbone of our base version and all other baselines\footnote{https://github.com/ymcui/LERT} for a fair comparison, the latest Chinese pre-trained language model that is trained on three types of linguistic features along with the original MLM pretraining task, bringing significant improvement over other variants \cite{chinese_bert}. We use AdamW as our optimizer and we set the initial learning rate as 5e-5 with a linear decay. In particular, we use sequences of 64 tokens for each utterance and 512 for each candidate, we set the minimum window size as 5 and the training batch size as 64, and train our models for 5 epochs using the combination of DuLeMon, KiDial, and Diamante datasets. For the evaluation, we consider three retrieval metrics: R@1, R@5, and MRR following \citet{polyencoder}. We set the size of the candidate pool as 64 during all evaluations without other statements.

\input{figs/zero_shot}

\subsection{Main Result}
With the setups described above, we fine-tuned the model on the datasets, and report the results in Table \ref{tab:main_exp}. It is exciting to see that UniversalCR$_{single}$ achieves better performance than all other baselines, particularly on knowledge selection (R@1 $\uparrow5.57\%$) and response selection tasks (R@1 $\uparrow4.27\%$), revealing the effectiveness of our framework. We also found that our method can not achieve consistent improvement at R@5, this is due to there being only one hard negative in the candidate pool for each query. We observed that the efficacy of our framework is positively correlated with the presence of more such difficult negative samples. In addition, we attribute the large improvement in knowledge and response selection tasks to the benefits of $\mathcal{L}_{pair}$ and $\mathcal{L}_{hist}$ since we model the more subtle order relationship between dialogue context and different candidates, which suit the design of corresponding datasets\footnote{since the knowledge and response selection is relatively sensitive with historical candidates compared with persona selection. For example, the candidates from the sample of KiDial likely come from the same document while sharing similar semantics.} and also downstream applications. Moreover, the performance of UniversalCR$_{full}$ is comparable to or even better than UniversalCR$_{single}$, which was already a highly effective model. These findings suggest a promising path towards building robust and omnipotent dialogue retrieval systems which learn to perform diversity retrieval tasks to complete the dialogue successfully.


\input{tables/ablation}

\section{Analysis}
In this section, we conduct detailed analysis experiments to demonstrate the superiority of our proposed method in various aspects: ablation study, zero-shot performance, and robustness.




\subsection{Ablation Study}
To test the effectiveness of different loss objectives, context-adaptive encoder, and multi-task learning, we conduct an ablation study by removing specific modules respectively and report the results on Table~\ref{tab:aba_study}. The performance of all candidate selection tasks drops when removing the context-adaptive encoder (\textit{PS: 28.73 $\rightarrow$ 21.10; KS: 86.67 $\rightarrow$ 83.64; RS: 59.94 $\rightarrow$ 53.68}) and drops to worst performance after removing $\mathcal{L}_{hist}$. We also find the model converges much faster with the help of $\mathcal{L}_{hist}$. Unsurprisingly, we observe a reduction when $\mathcal{L}_{pair}$ is removed. Thus we conclude that the combination of $\mathcal{L}_{pair}$ and $\mathcal{L}_{hist}$ leads to better performance and faster optimization while $\mathcal{L}_{pair}$ can well capture the order relationship between context and different candidates since it is a major driver of the performance gains achieved in knowledge selection (\textit{56.65 $\rightarrow$ 59.94}) and response selection task (\textit{85.27 $\rightarrow$ 86.67}) compared with persona selection task (\textit{28.52 $\rightarrow$ 28.73}).


\subsection{Zero-shot Performance}
To investigate the zero-shot performance of our proposed model, we use another three datasets to evaluate under two settings: zero-shot (aka, Zero-shot) and supervised fine-tuning (aka, SPT)\footnote{We keep the setting as same as the main experiment.}. The result can be found in Figure \ref{zero_shot}. Notably, we observe that UniversalCR$_{full}$ achieves higher performance on \textbf{all} three datasets than UniversalCR$_{single}$, especially at the DuSinc dataset which is even comparable with SPT method \footnote{We surprisingly find that the zero-shot performance of UniversalCR$_{full}$ on DuSinc is even higher than the SPT of other baselines, for example, 47.57 R@1/64 of UniversalCR$_{full}$ v.s. 40.33 R@1/64 of Bi-encoder}, although there still is a gap between SPT and Zero-shot setting. The disparity between the Zero-shot and SPT settings exhibits a considerable magnitude in the KdConv dataset no matter R@1 or R@5, which we attribute to the distinct design employed in our response selection dialogue dataset, namely Diamante \cite{diamante}, during the main experiment. Due to the involvement of human annotators in the process of selecting or amending model-generated candidate responses in a Diamante task, the approach to this task differs from the conventional response selection task, where negative responses are randomly selected. As a result, the transfer of knowledge from the diamante task to the conventional response selection task is limited without any fine-tuning. However, we argue the design of Diamante is much better and more realistic in practice with the development of LLMs \cite{llama}.

\input{tables/context_aba}
\subsection{The Effects of Previous Session}

In addition, we compare the performance of our proposed framework under the different choices of $K$ in Eq. \ref{k_ret} to investigate the influence of the previous session. We set the $K$ as [1,2,3,4] to retrieve $K$ most related utterance from the previous session and we also conduct an experiment in which we do not use any information from the previous session. To make a fair comparison, we evaluate the performance by loading the parameters from the latest three checkpoints and report the results in Figure \ref{window_size}. First of all, when comparing the performance of models that incorporate historical information with those that do not, it has been observed that the task of knowledge selection is particularly vulnerable and exhibits the greatest decrease in performance. This is hypothesized to stem from the fact that persona selection and response selection are comparatively more dependent on recent expressions, while knowledge selection differs in this regard. As conversations typically center around a specific topic, the inclusion of historical information can notably aid the model in effectively filtering out irrelevant information. Secondly, The model's performance is observed to degrade when the value of $K$ is either too small or too large in general. This is in agreement with the notion that an excessively small value of $K$ may result in important information being overlooked, while an overly large value may lead to the inclusion of noise data. Again, the knowledge selection task is more sensitive than the persona and response selection task with the choice of K. Due to these findings, we set the $K$ as 3 to get the best average performance during the main experiment.

\input{figs/window_size}

Besides that, we examined the effectiveness of a conventional dialogue processing method by directly concatenating all utterances together. However, we argue that this approach is impractical and inefficient for long dialogues in our setting. To provide a theoretical upper bound for our proposed method, we compared the performance of the two methods in different candidate selection tasks. The results, as presented in Table~\ref{tab:context_aba}, indicate that the gap in performance between the two methods is much larger in the response selection task compared to the other two tasks. We suggest that this may be due to the historical turns in the context providing additional information about how humans respond, enabling a better understanding of the response selection task. Overall, our findings suggest that the proposed method is more effective for long dialogues and can achieve better performance with the assistance of utterances from the previous session, even there is still a gap with the theoretical upper bound.

\subsection{The Effects of Different Sizes of Candidate Pool}
For the retrieval task, the size of the candidate pool is very important considering the efficiency of the retrieval model. We compared the performance of UniversalCR$_{full}$ under different sizes of the candidate pool (from 256 to 2) with a bi-encoder which is commonly used in large-scale retrieval tasks in Figure~\ref{pool_size}. Surprisingly, we find our model can achieve par with the bi-encoder even when the size is over 128, and our model additionally demonstrates undeniable and consistent improvement when the size is relatively small (less than 64) in all three tasks. These findings suggest that our model has promising potential to serve as both a retrieval and re-ranker model simultaneously, thanks to the introduction of pairwise similarity loss.

\input{figs/pool}

\section{Conclusion}
In this paper, we present a novel universal conversational retrieval framework, which can be applied to retrieve diverse external resources to complete the conversation successfully at the same time. We conduct extensive experiments on three major candidate selection tasks, including persona selection, knowledge selection, and response selection tasks. The experimental results suggest the effectiveness and potential of our framework to be a robust and omnipotent conversational retrieval system. In addition, we also found the framework demonstrates strong zero-shot performance and robustness serving as a re-ranker and a retriever simultaneously. We left other more complicated architectural improvements e.g. interactions between two encoder
towers in future work.


\section{Acknowledgement}

We would like to express our heartfelt gratitude to all anonymous reviewers for their insightful comments and suggestions. This research work is partially supported by CUHK direct grant No. 4055209 and CUHK Knowledge Transfer Project Fund No. KPF23GWP20.

\section{Bibliographical References}\label{sec:reference}

\bibliographystyle{lrec-coling2024-natbib}
\bibliography{lrec-coling2024-example}



\end{document}

%% file: figs/intro.tex
\begin{figure*}[t]
\centering
\includegraphics[trim={0cm 3cm 0cm 4cm}, clip, scale=1, width=1.0\textwidth]{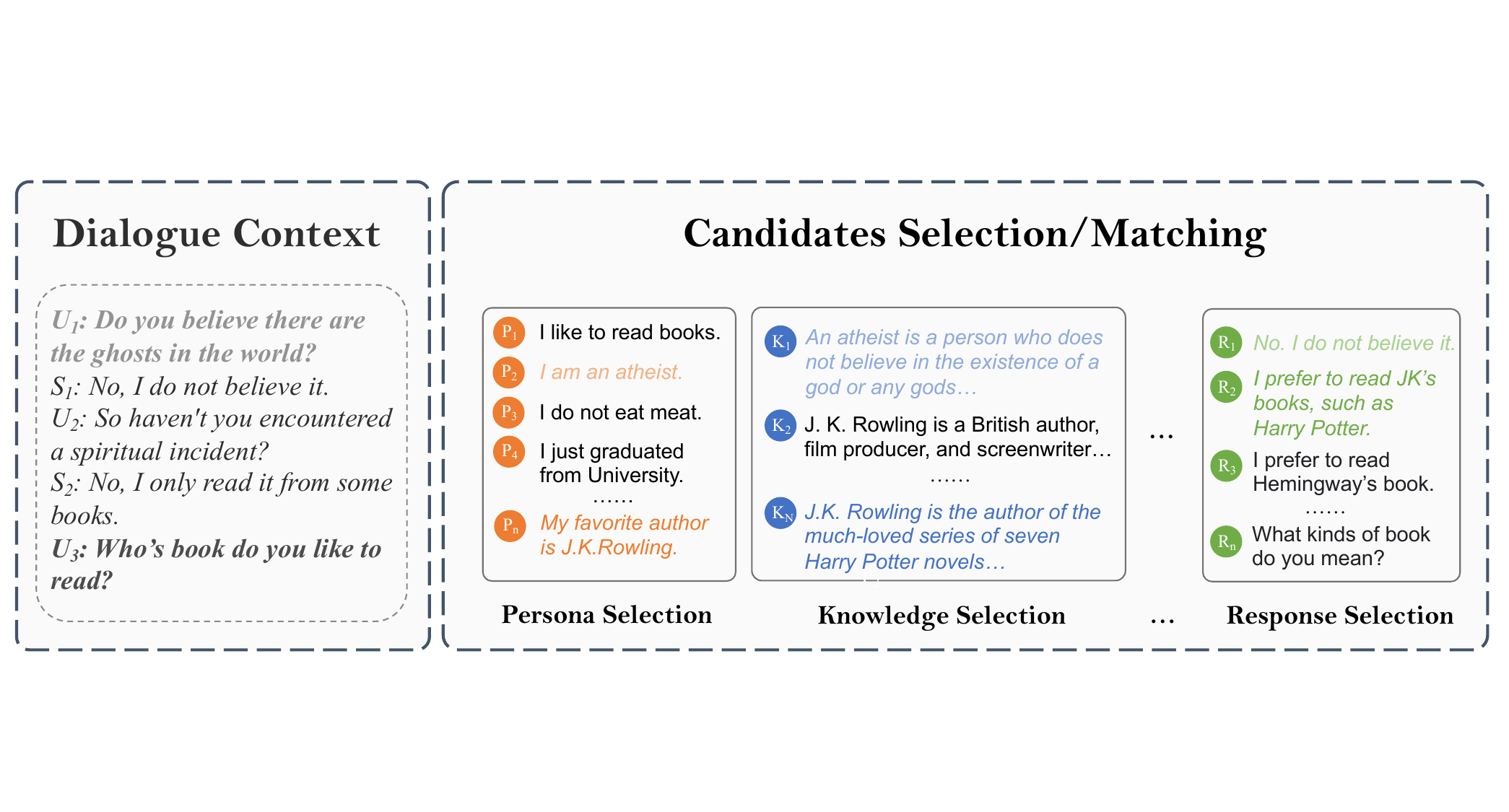}
\caption{Different candidates selection tasks in a dialogue system: persona selection, knowledge selection, and response selection task. According to $u_3$ in the dialogue context, it is obvious to select $p_n$, $k_n$, and $r_2$ as target persona, knowledge, and response for the next turn respectively, while the $p_2$, $k_1$, and $r_1$ are historical selected persona, knowledge and response for historical turn $u_1$. }
\label{intro_example}
\end{figure*}

%% file: figs/model.tex
\begin{figure}[t]
\centering
\includegraphics[trim={10cm 1cm 14cm 2cm}, clip, height=6cm]{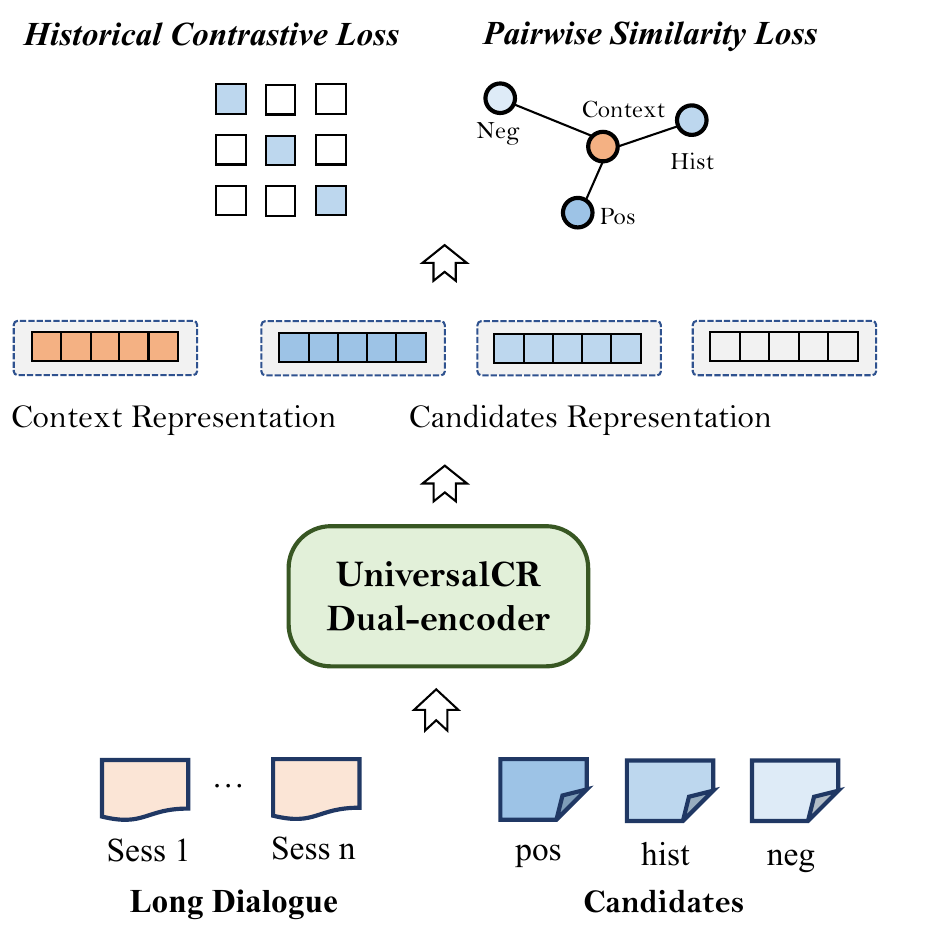}
\caption{The proposed Universal Conversational Retrieval based on Dual-encoder architecture, with the goal of optimizing \textit{historical contrastive loss} and \textit{pairwise similarity loss} thanks to the introduction of historical candidates. }
\label{model}
\end{figure}

%% file: tables/data_sts.tex
\begin{table}[t]
\centering
\fontsize{9}{11}\selectfont
\begin{tabular}{l|c|c|c|c}
\toprule
\textbf{Datasets} & \#Train & \#Dev & \#Test & \#All \\ \hline
DuLeMon & 28,243 & 1,993 & 2,036 & 30,202 \\
KBP & 4,788 & 589 & 584 & 5,961\\
\hline
Dusinc & 2,565 & 319 & 359 & 3,243 \\
KiDial & 21795 & 2813 & 2580 & 27188 \\
\hline
Diamante & 29,758 & 2,548 & 2,556 & 34,862\\
KdConv & 26,038 & 3,759 & 3,968 & 33,765 \\
\hline
All & 113187 & 12021 & 12083 & 137291 \\
\bottomrule
\end{tabular}
\vskip -0.225mm
\caption{The data statistics of used dialogue datasets, including persona-grounded dialogue dataset, knowledge-grounded dialogue dataset and also some conventional chit-chat dataset for response selection.}
\label{table:dataset}
\end{table}

%% file: tables/main_experiments.tex
\begin{table*}[!ht]
    \centering
    \begin{tabular}{l|ccc|ccc|ccc}
    \toprule
    \multirow{2}{*}{\textbf{Model}} & \multicolumn{3}{c|}{\textbf{Persona Sel.}} & \multicolumn{3}{c|}{\textbf{Knowledge Sel.}} & \multicolumn{3}{c}{\textbf{Response Sel.}}  \\
     \cline{2-10}    & R@1 & R@5 & MRR & R@1 & R@5 & MRR  & R@1 & R@5 & MRR \\
    \hline
    \hline
    BM25 & 0.06 & 0.38 & 9.49 & 0.35 & 1.45 & 28.37 & 0.59 & 1.35 & 30.72 \\ 
    DPR & 7.38 & 17.62 & - & 29.18 & 57.91 & - & 11.85 & 36.35 & - \\
    MultiCPR & 10.70 & 17.27 & - & 41.45 & 58.81 & - & 9.65 & 19.15 & - \\
    RocketQAv1 & 21.39 & 52.02 & 34.23 & 37.86 & 65.91 & 42.86 & 21.46 & 71.20 & 41.92 \\
    SentenceBERT & 18.99 & 47.64 & 32.91 & 43.57 & 86.59 & 61.13 & 35.45 & 73.59 & 51.86 \\
    Bi-Encoder &  26.79 & 56.13 & 40.46 & 80.08 & 98.88 & 86.14 & 52.93 & 90.73 & 68.69  \\
    Poly-Encoder 16 & 26.26 & 55.76 & 40.08 & 79.11 & 99.11 & 85.61 & 51.17 & \textbf{91.51} & 67.83 \\
    Poly-Encoder 32 & 25.73 & 55.76 & 39.80 & 78.76 & 98.91 & 85.46 & 50.67 & 90.88 & 67.49 \\
    RocketQAv2 & 21.87 & 50.80 & 34.27 & 34.62 & 61.64 & 39.71 & 21.87 & 70.23 & 42.31 \\
    \hline
    \hline
    UniversalCR$_{single}$ & 28.43 & 55.17 & 40.99 & 85.65 & 98.72 & 90.69 & 57.20 & 85.68 & 69.29 \\
    UniversalCR$_{full}$ & \textbf{28.73} & \textbf{55.99} & \textbf{41.47} & \textbf{86.67} & \textbf{99.22} & \textbf{91.45} & \textbf{59.94} & 87.09 & \textbf{71.73} \\
    \bottomrule
    \end{tabular}
    \caption{The performance of our proposed model and baselines on dataset DuLemon \cite{dulemon}, KiDial, and Diamante \cite{diamante}, correspond to persona selection, knowledge selection, and response selection. UnifiedD$_{sigle}$ simply fine-tune our model on each dataset instead of all in UnifiedD$_{full}$. }
    \label{tab:main_exp}
    \vspace{-5mm}
\end{table*}

%% file: figs/zero_shot.tex
\begin{figure*}[t]
\centering
\includegraphics[trim={0cm 0cm 0cm 0cm}, clip, scale=0.5, width=1.0\textwidth, height=5cm]{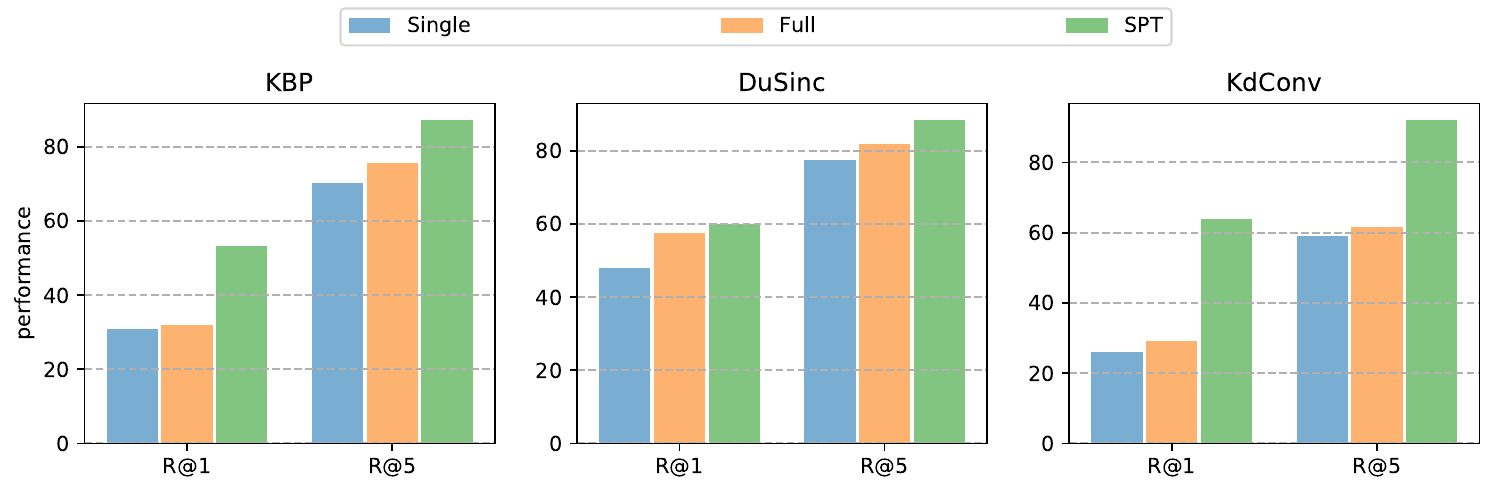}
\caption{The zero-shot performance of UnifiedD$_{single}$ and Unified$_{full}$, and the supervised fine-tuning result of UnifiedD$_{full}$ on three \textit{\textbf{New}} different datasets: Knowledge Behined Persona (persona selection), DuSinc (knowledge selection), and KdConv (response selection).}
\label{zero_shot}
\end{figure*}

%% file: tables/ablation.tex
\begin{table}[!t]
    \centering
    \begin{tabular}{l|ccc}
    \toprule
    \textbf{Model} & P.R@1 & K.R@1 & R.R@1 \\
    \hline
    \hline
    UniversalCR$_{full}$ & 28.73 & 86.67 & 59.94 \\
    -- \textit{context enc.} & 21.10 & 83.64 &53.68 \\
    -- $\mathcal{L}_{pair}$ & 28.52 & 85.27 & 56.65 \\
    -- $\mathcal{L}_{hist}$ & 20.14 & 43.64 & 34.12 \\
    \bottomrule
    \end{tabular}
    \caption{Ablation Study. The - \textit{context enc.} stands for considering all utterances in dialogue history by using a mean representation, and - $\mathcal{L}_{pair}$ and - $\mathcal{L}_{hist}$ means removing the corresponding loss constraint.}
    \label{tab:aba_study}
\end{table}

%% file: tables/context_aba.tex
\begin{table}[!t]
    \centering
    \begin{tabular}{l|ccc}
    \toprule
    \textbf{Model} & P.R@1 & K.R@1 & R.R@1 \\
    \hline
    \hline
    full concatenation &  31.41 & 89.34 & 65.34 \\
    context enc. & 28.73 & 86.67 & 59.94\\
    \bottomrule
    \end{tabular}
    \caption{The performance of different ways to process the dialogue history in which the full concatenation can be viewed as our theoretical upper bound.}
    \label{tab:context_aba}
\end{table}

%% file: figs/window_size.tex
\begin{figure}[!ht]
\centering
\includegraphics[trim={0cm 0cm 0cm 0cm}, clip, scale=1.0, width=0.45\textwidth, height=5cm]{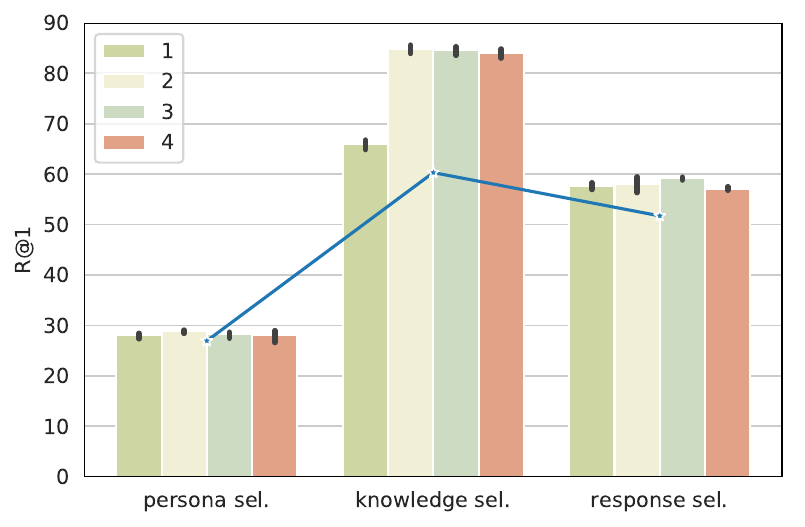}
\caption{The Performance of UnifiedD with different k or without any utterance from the previous session. \textcolor{blue}{Blue} line denotes the performance of Unified$_{full}$ without using any information from previous session. Here we report the R@1 metric.}
\label{window_size}
\end{figure}

%% file: figs/pool.tex
\begin{figure*}[ht]
\centering
\includegraphics[trim={0cm 0cm 0cm 0cm}, clip, scale=0.5, width=1.0\textwidth, height=5cm]{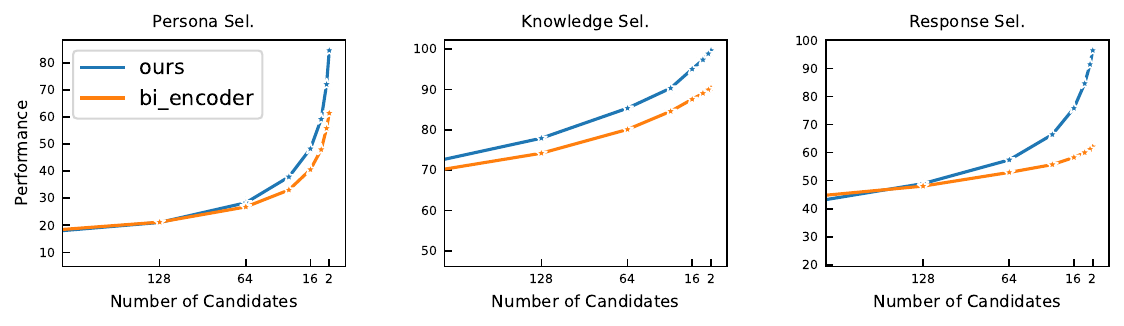}
\caption{The Performance of UniversalCR$_{full}$ on different selection tasks: persona selection, knowledge selection, and response selection, with the number of candidates ranging from 256 to 2.}
\label{pool_size}
\end{figure*}